# AutoRD: An Automatic and End-to-End System for Rare Disease Knowledge Graph Construction Based on Ontology-enhanced Large Language Models

Lang Cao[1], Jimeng Sun[1], Adam Cross[2]
[1] Department of Computer Science, University of Illinois Urbana-Champaign
[2] Department of Pediatrics, University of Illinois College of Medicine Peoria

## Abstract

**Background:** Rare diseases affect millions worldwide but often face limited research focus due to their low prevalence. This results in prolonged diagnoses and a lack of approved therapies. Recent advancements in Large Language Models (LLMs) have shown promise in automating the extraction of medical information, offering potential to improve medical diagnosis and management. However, most LLMs lack professional medical knowledge, especially concerning rare diseases, and struggle to handle the latest rare disease information. They also cannot effectively manage rare disease data and are not directly suitable for diagnosis and management tasks.
**Objective:** Our objective is to create an end-to-end system called AutoRD, which automates the extraction of information from medical texts about rare diseases, focusing on entities and their relations. AutoRD integrates up-to-date structured knowledge and demonstrates superior performance in rare disease extraction tasks. We conduct various experiments to evaluate AutoRD's performance, aiming to surpass common LLMs and traditional methods.
**Methods:** AutoRD is a pipeline system that involves data preprocessing, entity extraction, relation extraction, entity calibration, and knowledge graph construction. We implement this system using LLMs and medical knowledge graphs developed from open-source medical ontologies, utilizing techniques such as chain-of-thought reasoning and prompt engineering. We quantitatively evaluate our system's performance in entity extraction, relation extraction, and knowledge graph construction. The experiment uses the well-curated dataset RareDis2023, which contains medical literature focused on rare disease entities and their relations.
**Results:** On the RareDis2023 dataset, AutoRD achieves an overall entity extraction F1 score of 56.1% and a relation extraction F1 score of 38.6%, marking a 14.4% improvement over the baseline LLM. Notably, the F1 score for rare disease entity extraction reaches 83.5%, indicating high precision and recall in identifying rare disease mentions. These results demonstrate the effectiveness of integrating LLMs with medical ontologies in extracting complex rare disease information.
**Conclusions:** AutoRD is an automated end-to-end system for extracting rare disease information from text to build knowledge graphs, addressing critical limitations of existing LLMs. By leveraging ontology-enhanced LLMs, AutoRD constructs a robust medical knowledge base that incorporates up-to-date rare disease information, facilitating faster diagnoses and improved management of rare diseases. This work underscores the significant potential of LLMs in transforming healthcare, particularly in the rare disease domain. AutoRD not only advances rare disease information extraction but also paves the way for improved diagnoses and treatments, offering renewed hope to both patients and clinicians.
**Trial Registration:** N/A

## Introduction

### Objectives

Rare diseases, also known as orphan diseases, are relatively uncommon in isolation and sometimes receive less individual attention in medical research due to their low prevalence. [1] The likelihood of an individual being affected by a rare disease is relatively low. However, when considering the global population, many individuals are impacted. In the United States, rare diseases affect approximately 30 million people; [2] globally, the number rises to between 300 and 400 million. [**Error! Reference source not found.**] Furthermore, the rare disease patient population, distributed across 5,000 to 10,000 distinct diseases, [4] suffers from a significant lack of medical knowledge due to the rarity of a given illness. Consequently, patients often face prolonged and costly diagnostic processes and intensive treatments, with many of these diseases lacking approved therapies. [5,6] This situation underscores the substantial burden placed on both patients and healthcare systems. [7]

Online resources, including open-source databases, offer valuable references for medical professionals, contributing to the development of a comprehensive rare disease knowledge system. Examples of such databases include the Unified Medical Language System (UMLS) [8], the Human Phenotype Ontology (HPO) [9] and the Orphanet. [10] Specifically, Orphanet's database provides detailed information linking rare diseases, genes, and phenotypes, which greatly aids in the identification and diagnosis of rare diseases, among other related processes. However, these databases require considerable human effort for curation and maintenance. Therefore, there is an urgent need to develop methods that can support the process of establishing and enhancing rare disease medical knowledge systems.

Natural Language Processing (NLP) techniques are instrumental in automatically processing unstructured text to extract structured and clinically relevant information. This technique is especially beneficial for information extraction and knowledge discovery in the medical field. Recently, Large Language Models (LLMs) have demonstrated exceptional proficiency in language understanding and generation, garnering significant attention in the NLP domain. [11,12] Their ease of use allows humans to complete a wide range of complex tasks in everyday life. Moreover, the extensive knowledge stored within their parameters equips them to excel in domain-specific applications, such as medicine and healthcare. [13]

Current research is beginning to evaluate the capabilities of the most powerful LLMs, such as ChatGPT and GPT-4, across various medical applications. These applications include licensing examinations, [14] question answering, [15] and medical education. [16] Notably, several studies have demonstrated that LLMs are effective few-shot medical Named Entity Recognition (NER) extractors, exhibiting superior few-shot learning capabilities compared to other NLP methods. [17] In the context of rare diseases, where resources are often limited, LLMs emerge as valuable tools for extracting information about these conditions, showcasing their utility in enhancing medical knowledge systems.

In this paper we introduce Automated Rare Disease Mining (AutoRD) as an efficient tool for extracting information about rare diseases and constructing corresponding knowledge graphs. The system processes unstructured medical text as input and outputs extraction results and a knowledge graph. It is comprised of several key stages: data preprocessing, entity extraction, relation extraction, entity alignment, and knowledge graph construction. Among these, entity and relation extraction are the most critical parts. AutoRD is an LLM-based system built upon GPT-4. [12] We employ prompts as instructions to guide the LLMs through the entity and relation extraction processes. The model leverages its strong zero-shot capabilities to identify and extract entities and to analyze relationships between them. Although LLMs are pre-trained with extensive knowledge, they sometimes lack precise medical information. To address this, we enhance the LLM's medical knowledge using rare disease and phenotype ontologies. This is achieved by designing sophisticated prompts that incorporate relevant knowledge, including few-shot learning, structured output formats, and detailed guidance for LLMs. We conducted experiments to evaluate the system and identified the advantages and limitations of AutoRD. In summary, our contributions can be summarized as follows:

1. We propose AutoRD, an automated end-to-end system which efficiently extracts rare disease information from text and builds knowledge graphs. This is a useful and practical system which can help medical professionals discover information about rare diseases.
2. We use ontology-enhanced LLMs in the module of rare disease entity extraction and relation extraction. This approach harnesses the few-shot learning capabilities of LLMs and integrates medical knowledge from ontologies, resulting in an improved performance beyond what LLMs achieve alone.
3. We conduct experiments and provide extensive analysis to demonstrate the effectiveness of AutoRD on the carefully processed RareDis2023 dataset.

## Background and significance

Several studies have employed machine learning methods to support and enhance the medical management and process of rare diseases. Sanjak et al [18] introduced an innovative method for clustering over 3,000 rare diseases using node embeddings within a knowledge graph. This approach facilitates a deeper understanding of the relationships between different diseases and opens possibilities for drug repurposing. Alsentzer et al [19] developed Shepherd, a deep learning model designed for diagnosing rare diseases. This model effectively leverages clinical and genetic patient data along with existing medical knowledge to uncover new disease-gene associations. This work exemplifies the potential of AI in medical diagnostics, even in scenarios with limited labeled data. Rashid et al. [20] explored a unique approach in rare disease research through the National Mesothelioma Virtual Bank. They utilized REDCap and a web portal query tool to integrate and manage clinical data from multiple institutions. This method demonstrates the power of combining data management tools and web technologies to enhance research and collaboration in the field of rare diseases.

The advent of LLMs has led to their increasing application in the medical field. Datta et al [21] developed AutoCriteria, an LLM-based information extraction system that has shown high accuracy and generalizability in extracting detailed eligibility criteria from clinical trial documents for various diseases. This represents a scalable solution for clinical trial applications. In the context of rare diseases, there have been specific research efforts utilizing LLMs. Shyr et al [22] explored the performance of ChatGPT in extracting rare disease phenotypes from unstructured text, using zero- and few-shot learning techniques. This study demonstrated potential in certain scenarios, particularly with tailored prompts and minimal data. Oniani et al [23] proposed Models-Vote Prompting (MVP), an approach that improves LLM performance in few-shot learning scenarios by aggregating outputs from multiple LLMs through majority voting.

However, these studies on LLM applications for rare diseases are still preliminary. Both focused on evaluating the basic capabilities of LLMs in identifying rare diseases or used simple prompt ensembles to slightly enhance LLM performance. They did not explore the task of extracting relationships between rare diseases and related phenotypes. Additionally, these studies primarily explore basic applications of LLMs and do not extend to a comprehensive LLM-based system. Building on these foundational works, our research continues to delve into the use of LLMs for rare disease applications. Unlike prior efforts, we propose an integrated and useful system aimed at extracting rare disease information from unstructured text. The elaborate methods incorporated into this system significantly enhance extraction accuracy compared to the use of pure LLMs, marking a substantial advancement in this field.

## Methods

### Data

To improve the medical understanding of LLMs, we incorporated three medical ontologies into AutoRD: Orphanet Rare Disease Ontology (ORDO)**Error! Reference source not found.**, HPO-ORDO Ontology Module (HOOM) [10] , and Mondo Disease Ontology (Mondo) [24] .

For assessing the entity and relation extraction capabilities of AutoRD, the RareDis-v1 dataset [25] was employed. Prior to use, this dataset underwent several reprocessing steps including manual review and revision of annotation errors followed by reshuffling. We have named this new dataset RareDis2023.

### AutoRD Framework

We present AutoRD, an innovative system designed to automatically extract rare disease information from medical texts and create a knowledge graph. The AutoRD framework is illustrated in (Figure 1) and consists of a pipeline structure that includes data preprocessing, entity extraction, relation extraction, entity calibration, and knowledge graph construction. The extraction steps, which include entity and relation extraction, are the core components of the system. In these steps, we utilize large language models, along with medical ontologies, to effectively extract information from the texts.

Figure 1. The AutoRD framework processes medical texts as input data and outputs entities related to rare diseases and rare disease triples, which are the results of the extraction process. Subsequently, it constructs a knowledge graph based on these extraction results. During the entity and relation extraction steps, ontology-enhanced large language models are utilized to enhance performance.

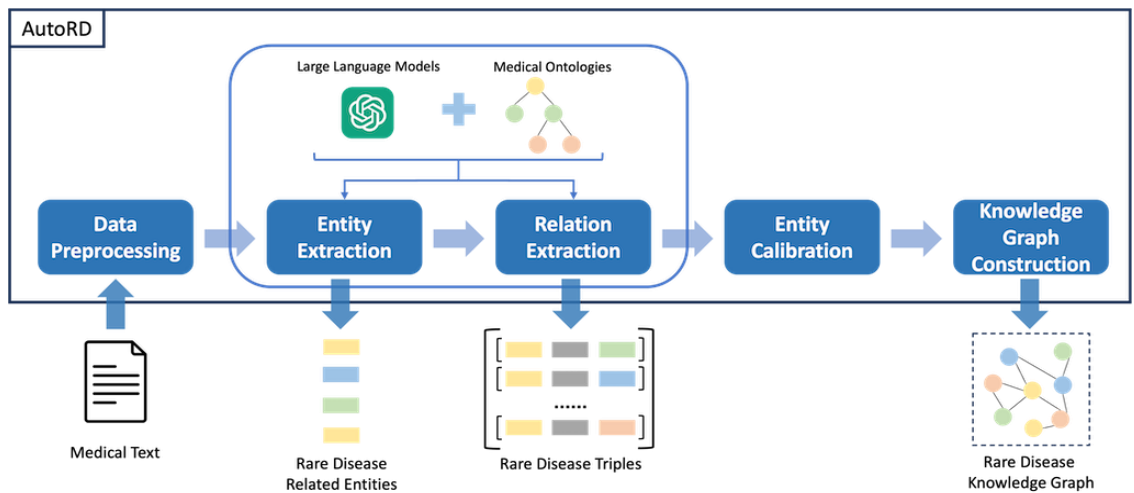

## Task Definition

Given a medical text T, AutoRD is designed to first extract entities E = {E1, E2, ..., En} and relations R = {R1, R2, ...., R}, and then output a knowledge graph KG based on E and R.

The medical text T can include clinical notes, research articles, or any text containing potential rare disease information. The final output, a knowledge graph KG, represents a set of rare diseases and related entities (such as diseases, symptoms, etc.) along with their relationships within the text.

The entity types listed in (Table 1) include 'rare_disease', 'disease', 'symptom_and_sign', and 'anaphor'. We group 'symptom' and 'sign' together because they both represent phenotypic abnormalities that may suggest a disease or medical condition. Distinguishing between them is not crucial in the context of rare disease research.

Table 1. Entity type in the entity extraction task. Definitions and examples of all entity types: 'rare_disease', 'disease', 'symptom_and_sign', and 'anaphor'. The definitions and examples are based on the original RareDis dataset definitions. [25]

| Entity Type | Definition | Example |
| --- | --- | --- |
| rare_disease | Diseases which affect a small number of people compared to the general population. A disease is often considered to be rare when it affects less than 1 in 2000 individuals. [26] | acquired aplastic anemia, Fryns syndrome, giant cell myocarditis |
| disease | An abnormal condition of a part, organ, or system of an organism resulting from various causes such as infection, inflammation, environmental factors, or genetic defect, and characterized by a patterned group of signs and/or symptoms. | cancer, Alzheimer, cardiovascular disease |
| symptom_and_sign | Signs and symptoms are abnormalities that may suggest a disease. A symptom is a physical or mental problem that a person experiences that may indicate a disease or condition; it is a subjective finding reported by the patient. In contrast, a sign is an observable or otherwise discoverable feature that is considered abnormal. | fatigue, dyspnea, pain inflammation, rash, abnormal heart rate, hypothermia |
| anaphor | Pronouns, words, or nominal phrases that refer to a rare disease (which is the antecedent of the anaphor) | This disease, These diseases |

Relation types are displayed in (Table 2), which include 'produces', 'increases_risk_of', 'is_a', 'is_acron', 'is_synon', and 'anaphora.' Each relation type represents a specific kind of relationship between a subject and an object, both of which can be any medical term entity.

Table 2. Relation types in the entity extraction task. Definitions of all relation types: 'produces', 'increases_risk_of', 'is_a', 'is_acron', 'is_synon', 'anaphora'. The definitions are based on those in the original RareDis dataset. [25]

| Relation Type | Definition |
|---|---|
| produces | Relation between any disease and a sign or a symptom produced by that disease. |
| increases_risk_of | Relation between a disease and a disorder, in which the presence of the disease increases the likelihood of the presence of the disorder. |
| is_a | Relation between a given disease and its classification as a more general disease. |
| is_acron | Relation between an acronym and its full or expanded form. |
| is_synon | Relation between two different names designating the same disease. |
| anaphora | Relation between an antecedent and an anaphor entity. The antecedent must be a rare disease. |

### Data Preprocessing

Before the system performs entity and relation extraction, we first preprocess the data due to the token limit of LLMs. In our system, we use GPT-4, which has a token limit of 8,000. Our maximum length for prompts is approximately 1,000 tokens, including the length of both input and output in the prompt slot. We divide lengthy input documents into segments containing fewer than 2,000 tokens to adhere to the token limit. To minimize entity relations across segments, we recognize that relations typically occur within a single natural paragraph. Therefore, we segment documents at natural paragraph boundaries, ensuring each segment contains fewer than 2,000 tokens. When paragraphs are segmented during preprocessing, some relations might span across the segmented parts, leading to incomplete extraction. To address this, we re-extract relations from the middle portions of previously segmented text to capture any new or missed relationships that may not have been fully identified in the initial extraction. This ensures that all relevant relationships within the text are accurately identified and included in the final knowledge graph.

We process medical knowledge data from ontology files downloaded from official websites. ORDO encompasses rare diseases that have been discovered up to the present day. From this ontology, we extract the names and definitions of all rare diseases. Mondo offers a unified medical terminology covering various medical concepts, from which we extract all disease, symptom, and sign concepts along with their definitions. Additionally, HOOM is an ontology that annotates the relationships between clinical entities and phenotypic abnormalities and reports their frequencies of occurrence. We extract information from HOOM as triples, consisting of (Rare Disease, Frequency, Phenotype). After preprocessing the ontology files, we can easily integrate medical knowledge from these ontologies into LLMs to enhance their knowledge base.

Data in the RareDis dataset also requires preprocessing for evaluation. The input texts in RareDis are all shorter than 512 tokens and consist of single paragraphs from medical literature, which contains a total of 1,040 data elements (texts and labels). We have corrected some errors in the annotations of the original dataset. To evaluate performance and compare it with the fine-tuning baseline, we divided the dataset into training, validation, and test sets in a 6:2:2 ratio, resulting in 624, 208, and 208 entries, respectively. The training set is used for training fine-tuning models and selecting some exemplars for LLMs, while the validation set is utilized to select the best fine-tuning models during training. In alignment with our task definition, we have merged 'Symptom' and 'Sign' from the original dataset into one entity type, 'symptom_and_sign'. We have named the newly processed dataset RareDis2023.

### Entity Extraction

After preprocessing, AutoRD subsequently carries out entity extraction. We drew inspiration from the concept of chain-of-thought (CoT) [27] to structure the entity extraction process. CoT proposes that tackling complex problems step by step can enhance the performance of LLMs. Similarly, we divided entity extraction into three sub-steps. In each step, an LLM completes a specific, smaller task. This division of the task allows us to integrate external medical knowledge more effectively from ontologies into the LLMs during this process. The three steps are: extract medical terms, extract more terms, and extract entities.

The first step, extract medical terms, extracts basic medical terms from the text. We only employ a string-match algorithm with negation detection in this process. A dictionary is built from the medical ontology Mondo. We use Mondo here because it encompasses nearly all standard medical terms. For each text, we use a string-match algorithm to search any medical terms in ontologies and save candidate medical terms as temporary results. For negation detection, we initially make a list of negation keywords manually, followed by the creation of a regular expression template. This template is then used to identify these keywords and extract the complete terms together. In this ontology, many terms include a free-text definition useful for model comprehension, so we make use of this information in subsequent steps.

The next step, extracting more terms, utilizes LLMs. The prompt template can be found in the left side of (Figure 2). We input the original text and medical terms extracted from the previous step into the LLMs. The LLMs then output additional medical terms. These include terms that are medically relevant but did not directly match an ontology term, including lemmatizations. This process leverages the strong language comprehension capabilities of LLMs for more flexible term extraction. In this step, LLMs also identify anaphors. In the prompt, we first outline the specifics of the current task and provide clear definitions of the entity types. Additionally, we include guidelines for the LLMs on identifying entities which are difficult to recognize. This part is significant and can be continuously improved by medical experts based on the performance of the LLMs and the results they produce. In many cases, LLMs may have misunderstandings in this task, so we need to use prompts to adjust for and correct their interpretations. Finally, we define the output format of the LLMs to be easily parsed, such as in JSON format.

Figure 2. Content of all prompt templates in AutoRD. This figure presents the simplified content of all prompts to provide a clear framework of the prompt structure. The black text represents the original text of the instructions. Grey text indicates a summary of each part of the instructions. Blue text highlights the prompt slots, where external information and inputs can be inserted.

The final step, entity extraction, also involves the use of LLMs. The instruction prompt template can be found in the central part of (Figure 2). The input for this step includes the medical terms and anaphors extracted during the previous step, while the output is comprised of all extracted entities with their appropriate categorizations. The framework of the prompt is formatted like earlier steps; however, additional external information is incorporated into the prompt slots, including medical terms, exemplars, anaphors, and rare disease knowledge. Here,

'rare disease knowledge' refers to terms that can be definitively classified as rare diseases, achieved by matching candidate entities with terms in ORDO.

Furthermore, we utilize the concept of in-context learning (ICL). [28] ICL employs exemplars to enhance the performance of LLMs. Each exemplar is a gold input-output pair, demonstrating the correct method of processing input and generating output for LLMs. This approach is beneficial for guiding the output format of LLMs and providing them with reference material and knowledge to inform their responses. Exemplars can be randomly selected from the training set. After completing these three steps, we can extract entities from text using LLMs.

### Relation Extraction

In our methodology, relation extraction is conducted after entity extraction. All identified entities are fed into LLMs, which then output the extracted relations. The prompt template used for instructing the LLMs is depicted in the central part of (Figure 2). The underlying logic of this process is akin to that of entity extraction. In the prompt, we initially provide an overview of the current task and establish clear definitions for both entity and relation types. We also include additional considerations for the LLMs to consider during relation extraction. Finally, we define the output format for the LLMs, which is structured to be easily parsed in JSON format. The prompt also contains examples of relation extraction.

For the extraction of rare disease knowledge, we utilize HOOM, an ontology which consists of rare disease-phenotype triples. This ontology provides information on symptoms and signs associated with rare diseases. We employ rare diseases as keys to construct a dictionary, enabling the identification of related triples through string matching. This external medical knowledge aids the LLMs in acquiring information about existing relationships between rare diseases and certain phenotypes.

### Entity Calibration

Our goal is to construct a knowledge graph based on the extraction results. We consider that many entities might not have defined relationships with other entities. Moreover, after analyzing the extraction results, we observed that entities without any relationships are more likely to be irrelevant or falsely ascribed as medical entities within the context. For instance, the system identifies the term 'disorder' during the entity extraction phase. However, in the relation extraction, the system fails to detect any 'anaphora' or other relations, indicating that it is merely a generic term and can be disregarded in this context. In other instances, some false symptoms and signs are also effectively eliminated. Therefore, we introduce entity calibration as an additional step after relation extraction. The prompt template for this task can be seen on the right side of (Figure 2). In this step, we provide all results obtained from the previous steps and use the LLMs to reanalyze the relationships, filtering out unrelated entities. By combining the results from both entity and relation extraction phases, we obtain the comprehensive outcome of the entire extraction process.

### Knowledge Graph Construction

After extracting entities and relations, we postprocess the data to prepare for knowledge graph construction. This includes aligning entities, which involves merging identical nodes in the knowledge graph. For each triple, we assess whether the subjects or objects are the same. We begin by converting the names to lowercase and then determining if they match. Additionally, we transform all anaphoric relations to their original names.

After postprocessing, we can easily construct the knowledge graph based on these triples. Specifically, we utilize Neo4j [29] for this purpose. Neo4j is a highly flexible and scalable graph database, designed to store and process complex networks of data. It enables efficient querying and management of interconnected information. Using the API of Neo4j, we add the rare disease triples into the graph database one by one. As a result, we can visualize our rare disease knowledge graph within the Neo4j platform.

### Evaluation

For the entity and relation extraction component, we quantitatively evaluate the performance of AutoRD using the processed RareDis2023 dataset.

Regarding our method, AutoRD, we specifically selected 'gpt-4-0613', a version of GPT-4 from OpenAI, for the LLMs. We set the LLM's temperature to 0 to ensure the most stable output. For each prediction with exemplars, we randomly choose 5 exemplars from the training set. The performance of AutoRD is evaluated exclusively on the test set, and detailed prompt templates are available in the source code.

To analyze the improvement our method brings compared to using only LLMs, we evaluate the performance of the base LLM. We use the same LLM, 'gpt-4-0613', and maintain all other settings identical to AutoRD. The prompts were developed collaboratively by clinicians and computer engineers. The detailed prompt template can be found in the source code.

For our fine-tuning model baseline, we selected BioClinicalBERT [30]. Entity recognition is performed through token classification based on BIO labels, and relationships are identified by concatenating the embeddings of two entities, followed by a linear classification. This model is trained on the training set, optimized according to the validation set, and finally evaluated on the test set. Detailed experimental settings are available in the source code.

In terms of evaluation metrics, we use Precision, Recall, and F1 metrics in a named entity recognition setting. For entity and relation extraction, we measure entity F1 and relation F1 respectively. The final overall results are represented by the overall F1 score, calculated as the mean of entity F1 and relation F1. We consider replicated entities in our extraction measurements, which are instances of the same entity occurring at different positions within the text. If the name of an extracted entity is correct, we regard it as true. The evaluation of relation extraction is not limited to correctly identified entities and for all true entities. We use the average score of entity F1 and relation F1 as the overall evaluation metric because both tasks are essential to AutoRD's performance. In some scenarios, simply identifying key rare disease entities is sufficient, while in others, understanding their relationships is equally important. By averaging these scores, we ensure a balanced assessment of the system's effectiveness across different scenarios.

In the test set, the entity instances include the follow number of each type: 463 'disease', 1054 'rare_disease', 1255 'symptoms_and_signs,' and 334 'anaphor'. Numbers of each type of relation include 1261 'produces', 62 'increases_risk_of', 188 'is_a', 55 'is_acron', 22 'is_synon', and 331 'anaphora'.

For the knowledge graph construction, we conduct qualitative experiments and provide examples of the knowledge graph results. In this case, we only use the extraction results from the test dataset of RareDis2023.

## Results

### Main Experimental Results

The primary experimental results are presented in (Table 3), which includes comparisons of entity and relation extraction performance among BioClinicalBERT (a fine-tuning model), Base GPT-4 (a base LLM), and AutoRD (our method). Overall, AutoRD achieves an overall F1 score of 47.3%. The system demonstrates superior performance over these two baselines, with an improvement of 0.8% in overall F1 score compared to the fine-tuning model and a 14.4% improvement compared to the base LLM. Recall is deemed more important than precision in this context because human effort can be used to validate extracted results. The primary goal is to extract all gold entities initially. In terms of recall, our overall recall improved by 18.4% compared to Base GPT-4 and by 6.6% compared to the fine-tuning models. For each extraction objective, AutoRD achieves an overall entity extraction F1 score of 56.1% ('rare_disease': 83.5%, disease: 35.8%, 'symptom_and_sign': 46.1%, 'anaphor': 67.5%) and an overall relation extraction F1 score of 38.6% ('produces': 34.7%, 'increases_risk_of': 12.4%, 'is_a': 37.4%, 'is_acronym': 44.1%, 'is_synonym': 16.3%, 'anaphora': 57.5%).

Table 3. The main experimental results of entity and relation extraction. Our methods surpass both the fine-tuning model (BioClinical-BERT) and the base LLM (Base GPT-4) in terms of overall F1 score (%).

| Method | | type | Precision | Recall | F1 |
|---|---|---|---|---|---|
| BioClinical BERT | Entity | rare_disease | 80.5 | 87.7 | 83.9 |
| | | disease | 53.2 | 46.0 | 49.3 |
| | | symptom_and_sign | 62.3 | 62.5 | 62.4 |
| | | anaphor | 89.9 | 93.7 | 91.7 |
| | | entity_overall | 70.9 | 72.0 | 71.4 |
| | Relation | produces | 49.7 | 13.6 | 21.4 |
| | | increases_risk_of | 0.0 | 0.0 | 0.0 |
| | | is_a | 80.0 | 4.3 | 8.1 |
| | | is_acron | 0.0 | 0.0 | 0.0 |
| | | is_synon | 0.0 | 0.0 | 0.0 |
| | | anaphora | 82.9 | 23.3 | 36.3 |
| | | relation_overall | 57.0 | 13.4 | 21.7 |
| | overall | | 64.0 | 42.7 | **46.5** |
| Base GPT4 | Entity | rare_disease | 94.8 | 38.4 | 54.7 |
| | | disease | 22.5 | 59.8 | 32.7 |
| | | symptom_and_sign | 48.7 | 41.7 | 44.9 |
| | | anaphor | 45.2 | 69.5 | 54.7 |
| | | entity_overall | 43.2 | 46.3 | 44.7 |
| | Relation | produces | 26.5 | 3.3 | 5.8 |
| | | increases_risk_of | 9.6 | 8.1 | 8.8 |
| | | is_a | 33.0 | 30.9 | 31.9 |
| | | is_acron | 17.1 | 21.8 | 19.2 |
| | | is_synon | 0.0 | 0.0 | 0.0 |
| | | anaphora | 41.7 | 55.3 | 47.6 |
| | | relation_overall | 32.5 | 15.6 | 21.1 |
| | overall | | 37.9 | 30.9 | **32.9** |
| AutoRD (Ours) | Entity | rare_disease | 93.1 | 75.6 | 83.5 |
| | | disease | 26.6 | 54.9 | 35.8 |
| | | symptom_and_sign | 45.8 | 46.5 | 46.1 |
| | | anaphor | 59.0 | 79.0 | 67.5 |
| | | entity_overall | 51.8 | 61.1 | 56.1 |
| | Relation | produces | 37.2 | 32.4 | 34.7 |
| | | increases_risk_of | 11.8 | 13.1 | 12.4 |
| | | is_a | 41.4 | 34.0 | 37.4 |
| | | is_acron | 49.2 | 40.0 | 44.1 |
| | | is_synon | 12.8 | 22.7 | 16.3 |
| | | anaphora | 52.4 | 63.7 | 57.5 |
| | | relation_overall | 39.8 | 37.5 | 38.6 |
| | overall | | 45.8 | 49.3 | **47.3** |

First, comparing Base GPT-4 with BioClinicalBERT reveals that BioClinicalBERT, with its ample training data, aligns well with the original dataset's distribution and excels in multiple metrics. However, LLMs were not trained to fit this distribution. This discrepancy leads to issues such as misclassifications and ambiguous entity boundaries in LLMs. In relation extraction, however, the fine-tuned model fails to detect relations like 'increases_risk_of', 'is_acron', and 'is_synon', whereas Base GPT-4 detects all but 'is_synon'. This demonstrates the strong zero-shot capabilities of LLMs, especially for relations sparsely represented in the training set.

When comparing AutoRD with BioClinicalBERT, it is apparent that while our method falls somewhat short in entity extraction, it excels in relation extraction. Specifically, the entity F1 score is 15.1% lower than this baseline, but the relation F1 score is 16.9% higher. This results in an overall performance that is 0.8% better. Relation extraction plays a pivotal role in the construction of knowledge graphs, as it is essential towards understanding the underlying relationships between entities. AutoRD leverages the few-shot learning capability of LLMs to better analyze relationships between medical entities. However, we observed a lower precision in AutoRD, primarily because it identifies too many entities as 'diseases' and sometimes misclassifies 'symptom_and_sign' according to its extraction results.

Furthermore, when comparing AutoRD with Base GPT-4, it is evident that our method significantly improves performance by 18.4%. The most notable improvement is a 37.2% increase in the recall of rare disease entities from Base GPT-4. Base GPT-4, with its poor analysis capability and lack of sufficient medical knowledge, struggles to identify all types of rare diseases. Overall, our approach demonstrates substantial improvements in most metrics.

**Ablation Study**
Table 4. The results of the ablation experiment. It clearly shows that each key component contributes to the improvement of AutoRD in terms of F1 score (%). The symbol (∇) represents the magnitude of the performance drop.

| Method | Entity F1 | ∇ | Relation F1 | ∇ | Overall F1 | ∇ |
|---|---|---|---|---|---|---|
| AutoRD (Ours) | **56.1** | - | **38.6** | - | **47.3** | - |
| AutoRD w/o Knowledge | 53.8 | -2.3 | 36.1 | -2.5 | 45.0 | -2.3 |
| AutoRD w/o Exemplars | 52.9 | -3.2 | 34.9 | -3.7 | 43.9 | -3.4 |
| AutoRD w/o Notice | 44.7 | -11.4 | 33.7 | -4.9 | 39.2 | -8.1 |

We conduct an ablation study to analyze the contribution of various components within AutoRD to the overall system. The results are presented in (Table 4), which clearly shows that each key component contributes to the improvement of AutoRD. Note that 'Knowledge' represents the external knowledge sourced from medical ontologies, while 'Notice' refers to the reminders for the LLMs. This study suggests that AutoRD can effectively utilize knowledge from both medical ontologies and exemplars. The 'Notice' component brings a significant improvement of 8.1% in overall F1. The current notices for LLMs in AutoRD have been carefully fine-tuned, demonstrating their effectiveness in adjusting and correcting for the LLMs' interpretations for extraction tasks.

### Error Analysis

We perform error analysis for each entity and relation type. To illustrate the distribution of extraction results, we use two confusion matrices: one for entity extraction results and another for relation extraction results. The results are shown in (Figure 3) and (Figure 4), respectively. The term 'Error' in the 'Predicted' axis refers to entities that have been incorrectly extracted, whereas 'Error' in the 'True' axis denotes real entities that were not extracted. We exclude replicated entities to simplify the computation of the confusion matrices.

In the entity extraction confusion matrix, there is significant confusion between the categories of 'disease' and 'rare_disease', possibly due to overlapping textual features. The 'symptom_and_sign' category exhibits high classification accuracy but is also prone to being misclassified as 'Error', suggesting the need for more distinctive features or additional contextual information in the dataset. The 'Anaphor' category was accurately classified with fewer errors, indicating that the system effectively captures its linguistic features. However, many predicted entities are incorrectly extracted and are labeled as 'Error', indicating that LLMs tend to extract more information with a low precision.

Figure 3. The confusion matrix of the entity extraction task results in RareDis2023.

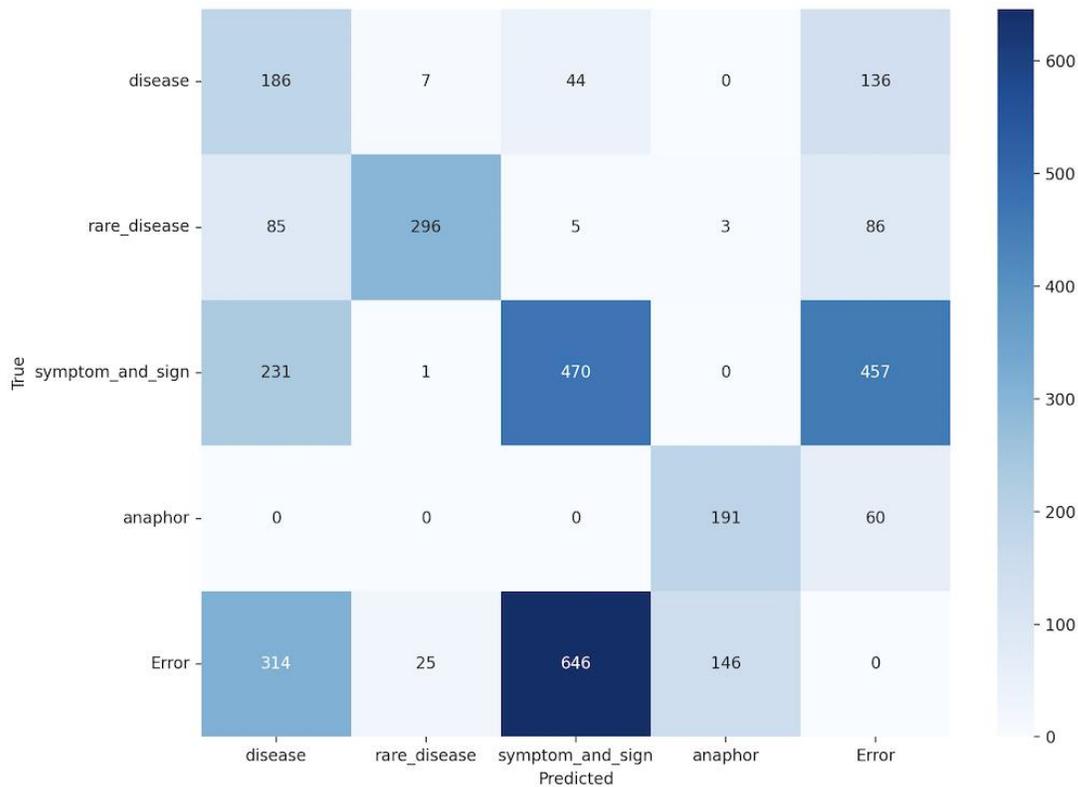

The confusion matrix for the relation extraction indicates varying degrees of performance across different categories. The 'produces' relation is often identified correctly but also often misclassified as 'Error', indicating recognition issues. 'Increases_risk_of' is more frequently an 'Error' than correct, demonstrating the recognition difficulty of this relation. 'Is_a' has moderate success but high error rates as well. 'Is_acron' and 'is_synon' rarely hit true positives and mostly fall into 'Error', possibly due to acronym variability and synonym recognition failure. 'Anaphora' resolution is relatively accurate but also misclassified, hinting at context comprehension challenges. The 'Error' category's high rate of both true and false positives is primarily affected by the false results from entity extraction step before it.

Figure 4. The confusion matrix of the relation extraction task results in RareDis2023.

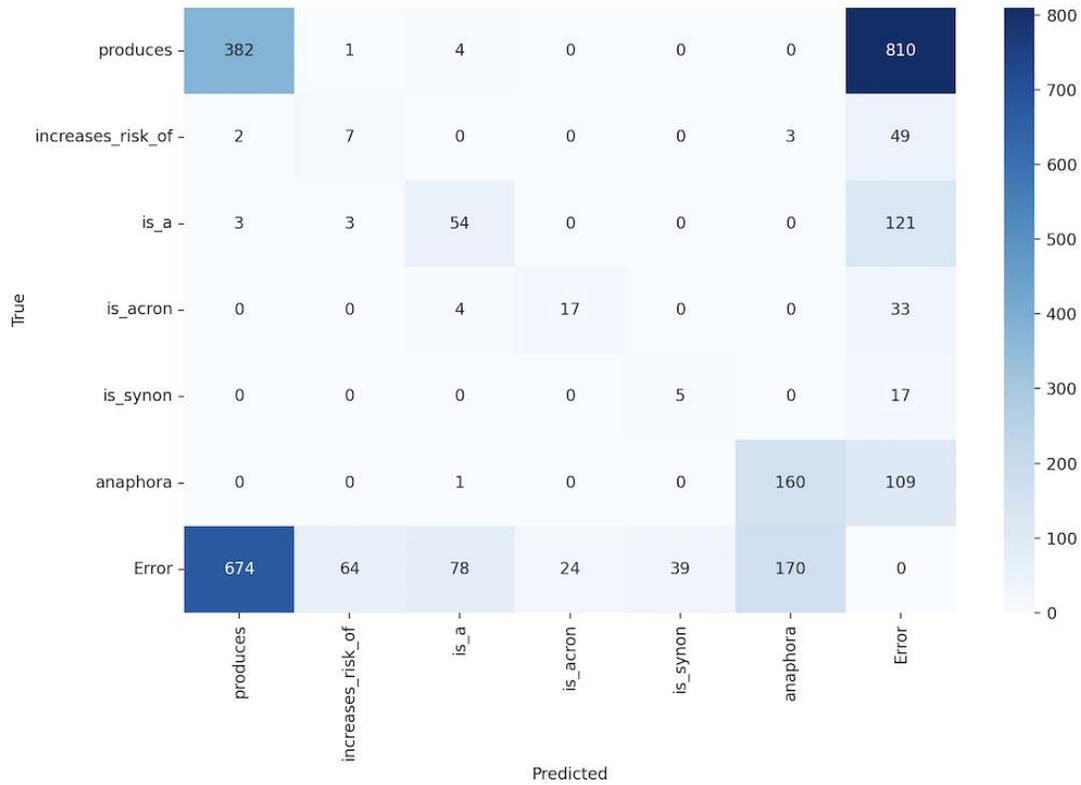

## Qualitative Results

Qualitative results are showcased in (Figure 5), which depicts all the extracted results from the RareDis2023 dataset. Our qualitative results have been validated by medical experts and have shown satisfactory outcomes. This visualization provides a global perspective, highlighting the relationships among various rare diseases and their associated signs and symptoms in a concise knowledge graph.

Figure 5. The example of constructed knowledge graph from RareDis2023. The result is a clear and well-structured knowledge graph.

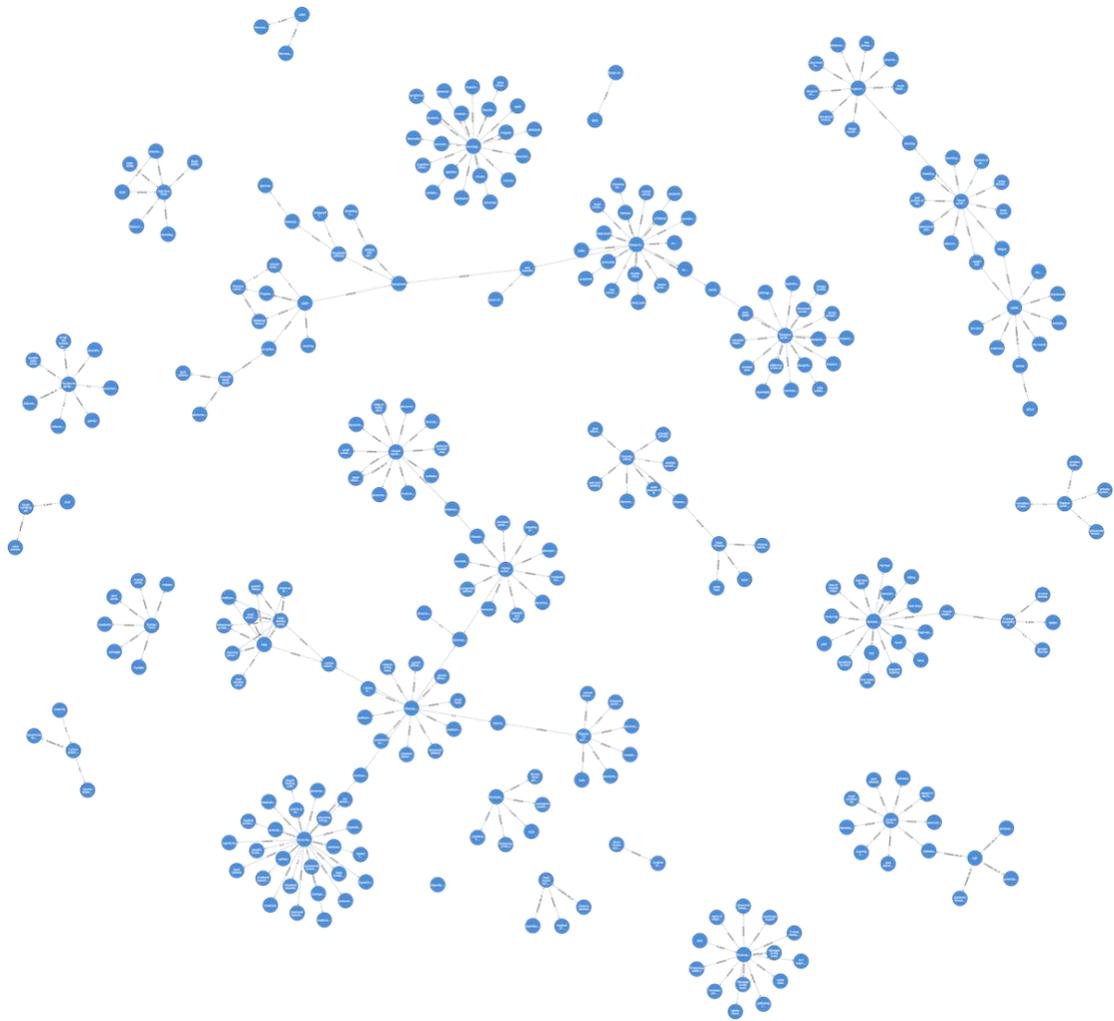

Specific extraction results of the knowledge graph are depicted in (Figure 6). This figure offers visualizations from a local perspective, illustrating an ideal structure of the knowledge graph. In this structure, rare diseases are positioned at the center of radial formations, with connections extending to entities like symptoms and signs. For example, in
, the rare disease 'Turcot syndrome' is associated with 'abdominal pain', 'bleeding', 'fatigue', etc.

Figure 6. An example provides a detailed view of a specific local section of the constructed knowledge graph.

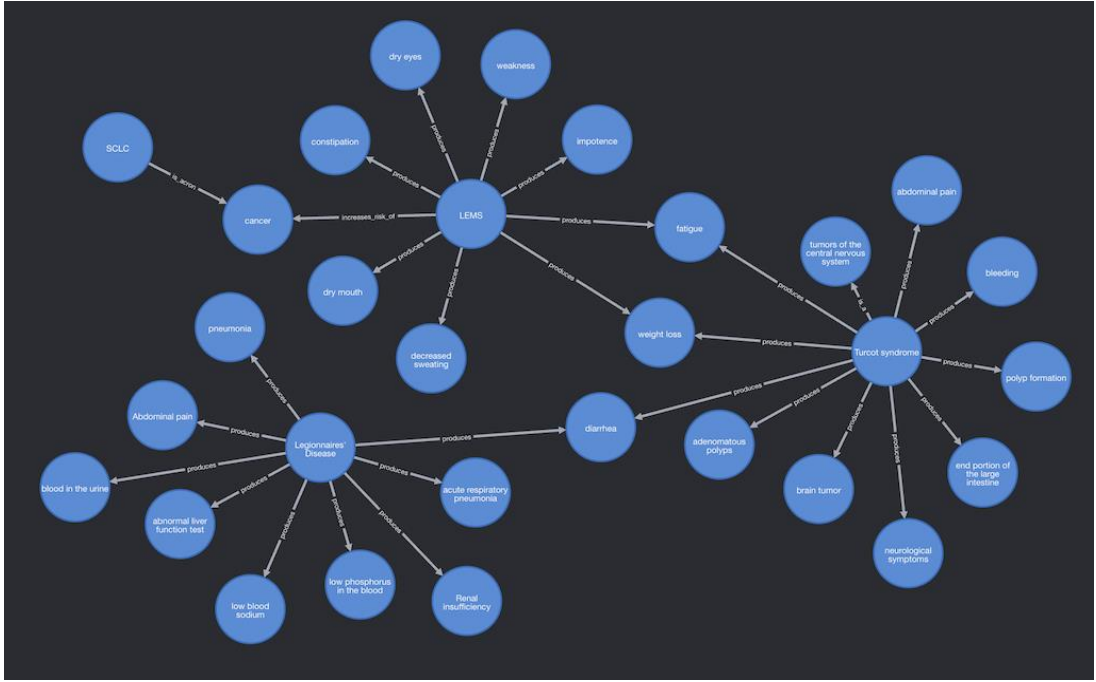

Additionally, we experimented with training specialized medical LLMs and compared their performance. Specifically, we utilized Camel-Platypus2-70B [31], a healthcare-tailored model that is an extension of Llama-2 [32], through continuous training. Our experiments revealed that, without specific training, this type of model struggles to execute complex tasks such as joint entity and relation extraction. It appears that the inherent medical knowledge is not readily applicable in these scenarios.

## Discussion

### Principal Results

Our experimentation demonstrates the effectiveness of our proposed system, AutoRD. It significantly improves upon the base LLM and even outperforms fine-tuning models without requiring any training. Within several designs, the incorporation of medical ontologies has notably enhanced the LLMs by addressing gaps in medical knowledge. Furthermore, the results achieved in knowledge graph construction by our system are commendable. We highlight the advantage of LLMs in low-resource scenarios like rare disease extraction, showcasing their vast potential. Our meticulously designed system, AutoRD, substantiates this claim. The emergence of LLMs is generating unparalleled opportunities in the phenotyping of rare diseases. These models facilitate the automatic identification and extraction of concepts related to these diseases. Our prompts are easily adjustable due to their clear structure, allowing for simple modifications. Additionally, medical knowledge derived from external sources can be updated at any time within the AutoRD system.

## Limitations

Nevertheless, there is considerable potential for further improvement with respect to AutoRD. For instance, integrating advanced text processing tools and specialized medical tools into our system could amplify its capabilities. In the future, we can deploy more powerful medical LLMs as base models to enhance medical understanding. Moreover, medical experts can contribute more tailored prompts to improve LLMs' performance.

Our work has potential limitations and avenues for extension. For example, we have only evaluated AutoRD on a single dataset, so the results may not fully reflect the system's performance across the entire spectrum of rare diseases or in other long-text scenarios. Additionally, the prompts we designed are intuitive, but there is still room for continuous tuning and experimentation of different prompts. We acknowledge that AutoRD may not be the optimal LLM application for this task, yet it significantly improves upon the baseline performance of LLMs. This work aims at demonstrating the potential of LLM applications in the healthcare field.

## Conclusions

AutoRD represents a significant advancement in the extraction of rare disease information, directly addressing the critical gaps associated with common LLMs used in rare disease medical research. By streamlining the process of building comprehensive knowledge graphs from unstructured medical texts, AutoRD tackles the substantial burden placed on patients and healthcare systems due to prolonged and costly diagnostic processes associated with rare diseases. By integrating ontology-enhanced large language models (LLMs), AutoRD overcomes the limitations of existing systems—particularly the significant human effort required for curation and maintenance of rare disease databases and the inability to handle complex and up-to-date rare disease information effectively.

Our experimental results demonstrate the system's effectiveness, achieving a 14.4% improvement over the most advanced LLM in both entity and relation extraction F1 scores. This enhancement effectively fills critical gaps in rare disease research by providing an automated method to support the establishment and enhancement of rare disease medical knowledge systems. By leveraging LLMs' strong zero-shot capabilities and integrating medical knowledge from ontologies, AutoRD contributes to a more robust and comprehensive medical knowledge base, ultimately facilitating faster diagnoses and improved management of rare diseases.

This study highlights AutoRD's potential to transform rare disease diagnostics and treatment by offering a scalable, automated solution for medical information extraction. The enhanced precision and recall in identifying rare disease entities and their relationships provide valuable insights for healthcare professionals, ultimately supporting better clinical decision-making and improved patient outcomes.

Furthermore, AutoRD's flexible architecture—incorporating techniques such as chain-of-thought and prompt engineering—offers promising opportunities for adaptation in other healthcare domains, especially in low-resource environments where medical expertise may be scarce.

In conclusion, AutoRD not only elevates the accuracy and efficiency of rare disease information extraction but also paves the way for future applications in medical diagnostics and personalized healthcare. By bridging the gap between vast unstructured medical data and actionable knowledge, AutoRD stands to significantly impact the fight against rare diseases, offering renewed hope to patients and clinicians alike as we move toward a future where advanced AI technologies play a central role in healthcare innovation.

### Acknowledgements
Lang Cao designed the main method, wrote codes, conducted experiments, and wrote the manuscript. Dr. Adam Cross and Professor Jimeng Sun participated in the design, analysis, and discussion of the experiment and helped to improve the method. Dr. Adam Cross and Professor Jimeng Sun also assisted with revisions to the manuscript.

### Conflicts of Interest
none declared.

### Abbreviations
AutoRD: Automatic Rare Disease Mining System
CoT: chain-of-thought
GPT: Generative Pre-trained Transformer
HOOM: HPO-ORDO Ontology Module
HPO: Human Phenotype Ontology
LLM: Large Language Model
UMLS: Unified Medical Language
Mondo: Mondo Disease Ontology
NER: Named Entity Recognition
NLP: Natural Language Processing
ORDO: Orphanet Rare Disease Ontology

### Multimedia Appendix 1
N/A